\documentclass[10pt,twocolumn, letterpaper]{article}

\usepackage{iccv}
\usepackage{times}
\usepackage{epsfig}
\usepackage{graphicx}
\usepackage{amsmath}
\usepackage{amssymb}
\usepackage{kotex}
\usepackage{booktabs}
\usepackage{multirow}
\usepackage{dsfont}
\usepackage{xcolor}
\usepackage{subcaption}
\usepackage{pifont}
\usepackage{algorithm,algorithmicx,algpseudocode}
\usepackage[pagebackref=true,breaklinks=true,letterpaper=true,colorlinks,bookmarks=false]{hyperref}

\iccvfinalcopy

\definecolor{darkergreen}{RGB}{21, 152, 56}
\definecolor{red2}{RGB}{252, 54, 65}
\newcommand\redp[1]{\textcolor{red2}{(#1)}}
\newcommand\greenp[1]{\textcolor{darkergreen}{(#1)}}

\ificcvfinal\fi
\begin{document}

\title{CutMix: Regularization Strategy to Train Strong Classifiers \\ with Localizable Features}

\author{
Sangdoo Yun$^1$~~~~~~~~~Dongyoon Han$^1$~~~~~~~~~Seong Joon Oh$^2$~~~~~~~~~Sanghyuk Chun$^1$\\ 
Junsuk Choe$^{1,3}$~~~~~~~~Youngjoon Yoo$^1$\\ 
~\\
$^1$Clova AI Research, NAVER Corp.\\
$^2$Clova AI Research, LINE Plus Corp.\\
$^3$Yonsei University \\
}

\maketitle

\begin{abstract}

Regional dropout strategies have been proposed to enhance the performance of convolutional neural network classifiers. They have proved to be effective for guiding the model to attend on less discriminative parts of objects (\eg leg as opposed to head of a person), thereby letting the network generalize better and have better object localization capabilities.
On the other hand, current methods for regional dropout remove informative pixels on training images by overlaying a patch of either black pixels or random noise. 
{Such removal is not desirable because it leads to information loss and inefficiency during training.}
We therefore propose the \textbf{CutMix} augmentation strategy: patches are cut and pasted among training images where the ground truth labels are also mixed proportionally to the area of the patches.
By making efficient use of training pixels and \mbox{retaining} the regularization effect of regional dropout, CutMix consistently outperforms the state-of-the-art augmentation strategies on CIFAR and ImageNet classification tasks, as well as on the ImageNet weakly-supervised localization task.
Moreover, unlike previous augmentation methods, our CutMix-trained ImageNet classifier, when used as a pretrained model, results in consistent performance gains in Pascal detection and MS-COCO image captioning benchmarks.
We also show that CutMix improves the model robustness against input corruptions and its out-of-distribution detection performances.
Source code and pretrained models are available at \href{https://github.com/clovaai/CutMix-PyTorch}{https://github.com/clovaai/CutMix-PyTorch}.
\end{abstract}

\section{Introduction}

Deep convolutional neural networks (CNNs) have shown promising performances on various computer vision problems such as image classification~\cite{ImageNet,alexnet,resnet}, object detection~\cite{fasterrcnn,liu2016ssd}, semantic segmentation~\cite{chen2018deeplab,Long_2015_CVPR}, and video analysis~\cite{nam2016learning,simonyan2014two}. 
To further improve the training efficiency and performance, a number of training strategies have been proposed, including data augmentation~\cite{alexnet} and regularization {techniques}~\cite{Dropout,stochasticdepth,szegedy2016rethinking_labelsm}.

In particular, to prevent a CNN from focusing too much on a small set of intermediate activations or on a small region on input images, random feature removal regularizations have been proposed. Examples include dropout~\cite{Dropout} for randomly dropping hidden activations and regional dropout~\cite{devries2017cutout,zhong2017randomerase,singh2017hide,ghiasi2018dropblock,choe2019attention} for erasing random regions on the input.
Researchers have shown that the feature removal strategies improve generalization and localization by letting a model attend not only to the most discriminative parts of objects, but rather to the entire object region~\cite{singh2017hide,ghiasi2018dropblock}.

\begin{table}[t]
\small
\tabcolsep=0.07cm
\begin{tabular}{@{}ccccc@{}}
 & ResNet-50  &  Mixup~\cite{zhang2017mixup} & Cutout~\cite{devries2017cutout} & CutMix \\
\multirow{4}{*}{Image} &  \multirow{4}{*}{\includegraphics[width=0.185\linewidth]{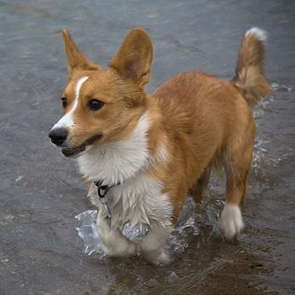}}
 &  \multirow{4}{*}{\includegraphics[width=0.185\linewidth]{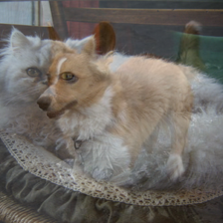}}
 &  \multirow{4}{*}{\includegraphics[width=0.185\linewidth]{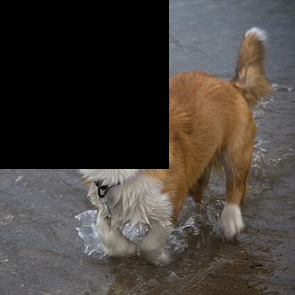}}
 &  \multirow{4}{*}{\includegraphics[width=0.185\linewidth]{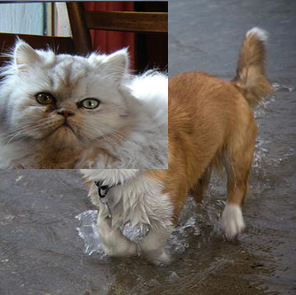}} \\ 
 & & & & \\ 
 & & & & \\
 & & & & \\ \midrule
\multirow{2}{*}{Label} 
&\multirow{2}{*}{Dog 1.0}
&\multirow{2}{*}{\begin{tabular}[c]{@{}c@{}}Dog 0.5\\ Cat 0.5\end{tabular}}
&\multirow{2}{*}{Dog 1.0}
&\multirow{2}{*}{\begin{tabular}[c]{@{}c@{}}Dog 0.6\\ Cat 0.4\end{tabular}} \\ 
  & & & & \\  \midrule
\multirow{2}{*}{\begin{tabular}[c]{@{}c@{}}ImageNet\\ Cls (\%) \end{tabular}}
& \multirow{2}{*}{\begin{tabular}[c]{@{}c@{}}76.3\\ (+0.0) \end{tabular}} 
& \multirow{2}{*}{\begin{tabular}[c]{@{}c@{}}77.4\\ \greenp{+1.1} \end{tabular} } 
& \multirow{2}{*}{\begin{tabular}[c]{@{}c@{}}77.1\\ \greenp{+0.8} \end{tabular}} 
& \multirow{2}{*}{\begin{tabular}[c]{@{}c@{}}\textbf{78.6}\\ \textbf{\greenp{+2.3}} \end{tabular}} \\
  & & & & \\  \midrule
\multirow{2}{*}{\begin{tabular}[c]{@{}c@{}}ImageNet\\ Loc (\%) \end{tabular}}
& \multirow{2}{*}{\begin{tabular}[c]{@{}c@{}}46.3\\ (+0.0) \end{tabular}} 
& \multirow{2}{*}{\begin{tabular}[c]{@{}c@{}}45.8\\ \redp{-0.5} \end{tabular} } 
& \multirow{2}{*}{\begin{tabular}[c]{@{}c@{}}46.7\\ \greenp{+0.4} \end{tabular}} 
& \multirow{2}{*}{\begin{tabular}[c]{@{}c@{}}\textbf{47.3}\\ \textbf{\greenp{+1.0}} \end{tabular}} \\
  & & & & \\ \midrule
\multirow{2}{*}{\begin{tabular}[c]{@{}c@{}}Pascal VOC\\ Det (mAP) \end{tabular}}
& \multirow{2}{*}{\begin{tabular}[c]{@{}c@{}}75.6\\ (+0.0) \end{tabular}} 
& \multirow{2}{*}{\begin{tabular}[c]{@{}c@{}}73.9\\ \redp{-1.7} \end{tabular} } 
& \multirow{2}{*}{\begin{tabular}[c]{@{}c@{}}75.1\\ \redp{-0.5} \end{tabular}} 
& \multirow{2}{*}{\begin{tabular}[c]{@{}c@{}}\textbf{76.7}\\ \textbf{\greenp{+1.1}} \end{tabular}} \\
  & & & & \\ \toprule
\end{tabular}
\caption{Overview of the results of Mixup, Cutout, and our CutMix on ImageNet classification, ImageNet localization, and Pascal VOC 07 detection (transfer learning with SSD~\cite{liu2016ssd} finetuning) tasks.
     Note that CutMix significantly improves the performance on various tasks.
     }
\label{tab:teaser}
\end{table}

While regional dropout strategies have shown improvements of classification and localization performances to a certain degree, deleted regions are usually zeroed-out~\cite{devries2017cutout,singh2017hide} or filled with random noise~\cite{zhong2017randomerase}, greatly reducing the proportion of informative pixels on training images.
We recognize this as a severe conceptual limitation as CNNs are generally data hungry~\cite{InstagramNet}.
How can we maximally utilize the deleted regions, while taking advantage of better generalization and localization using regional dropout?

We address the above question by introducing an augmentation strategy \textbf{CutMix}. 
Instead of simply removing pixels, we replace the removed regions with a patch from another image (See Table~\ref{tab:teaser}).
The ground truth labels are also mixed proportionally to the number of pixels of combined images.
{CutMix now enjoys the property that there is no uninformative pixel during training, making training efficient, while retaining the advantages of regional dropout to attend to non-discriminative parts of objects.}
The added patches further enhance localization ability by requiring the model to identify the object from a partial view.
The training and inference budgets remain the same.

CutMix shares similarity with Mixup~\cite{zhang2017mixup} which mixes two samples by interpolating both the image and labels.  
{While certainly improving classification performance, Mixup samples tend to be unnatural (See the mixed image in Table~\ref{tab:teaser}).
CutMix overcomes the problem by replacing the image region with a patch from another training image.}

Table~\ref{tab:teaser} gives an overview of Mixup~\cite{zhang2017mixup}, Cutout~\cite{devries2017cutout}, and CutMix on image classification, weakly supervised localization, and transfer learning to object detection methods.
Although Mixup and Cutout enhance ImageNet classification, they decrease the ImageNet localization or object detection performances. 
On the other hand, CutMix consistently achieves significant enhancements across three tasks.

We present extensive evaluations of CutMix on various CNN architectures, datasets, and tasks. 
Summarizing the key results, CutMix has significantly improved the accuracy of a baseline classifier on CIFAR-100 and has obtained the state-of-the-art top-1 error $14.47\%$. On ImageNet~\cite{ImageNet}, applying CutMix to ResNet-50 and ResNet-101~\cite{resnet} has improved the classification accuracy by $+2.28\%$ and $+1.70\%$, respectively. 
On the localization front, CutMix improves the performance of the weakly-supervised object localization (WSOL) task on CUB200-2011~\cite{wah2011cub} and ImageNet~\cite{ImageNet} by {$+5.4\%$} and $+0.9\%$, respectively. 
The superior localization capability is further evidenced by fine-tuning a detector and an image caption generator on CutMix-ImageNet-pretrained models; the CutMix pretraining has improved the overall detection performances on Pascal VOC~\cite{pascalvoc} by $+1$ mAP and image captioning performance on MS-COCO~\cite{lin2014microsoft} by $+2$ BLEU scores.
{CutMix also enhances the model robustness and alleviates the over-confidence issue \cite{hendrycks2016baseline, liang2017odin} of deep networks.}

\section{Related Works}

\noindent\textbf{Regional dropout: }
Methods~\cite{devries2017cutout,zhong2017randomerase} removing random regions in images have been proposed to enhance the generalization performance of CNNs. 
Object localization methods~\cite{singh2017hide,choe2019attention} also utilize the regional dropout techniques for improving the localization ability of CNNs.
CutMix is similar to those methods, while the critical difference is that the removed regions are filled with patches from another training images.
DropBlock~\cite{ghiasi2018dropblock} has generalized the regional dropout to the feature space and have shown enhanced generalizability as well. 
CutMix can also be performed on the feature space, as we will see in the experiments.

\noindent\textbf{Synthesizing training data: }
Some works have explored synthesizing training data for further generalizability.
Generating new training samples by Stylizing ImageNet~\cite{ImageNet,geirhos2018imagenet_stylize} has guided the model to focus more on shape than texture, leading to better classification and object detection performances. 
CutMix also generates new samples by cutting and pasting patches within mini-batches, leading to performance boosts in many computer vision tasks; unlike stylization as in~\cite{geirhos2018imagenet_stylize}, CutMix incurs only negligible additional cost for training.
For object detection, object insertion methods~\cite{dwibedi2017cut_paste_learn, dvornik2018modeling} have been proposed as a way to synthesize objects in the background.
These methods aim to train a good represent of a single object samples, while CutMix generates combined samples which may contain multiple objects.

\noindent\textbf{Mixup: }
CutMix shares similarity with Mixup~\cite{zhang2017mixup, tokozume2018betweenclass} in that both combines two samples, where the ground truth label of the new sample is given by the linear interpolation of one-hot labels. 
As we will see in the experiments, Mixup samples suffer from the fact that they are locally ambiguous and {unnatural}, and therefore confuses the model, especially for localization.
Recently, Mixup variants~\cite{verma2018manifoldmixup, summers2019improvedm, guo2018adamix, takahashi2018ricap} have been proposed; they perform feature-level interpolation and other types of transformations.
Above works, however, generally lack a deep analysis in particular on the localization ability and transfer-learning performances.
We have verified the benefits of CutMix not only for an image classification task, but over a wide set of localization tasks and transfer learning experiments. 

\noindent\textbf{Tricks for training deep networks: }
Efficient training of deep networks is one of the most important problems in computer vision community, as they require great amount of compute and data.
Methods such as weight decay, dropout~\cite{Dropout}, and Batch Normalization~\cite{BN} are widely used to efficiently train deep networks. 
Recently, methods adding noises to the internal features of CNNs ~\cite{stochasticdepth,ghiasi2018dropblock,yamada2018shakedrop} or adding extra path to the architecture~\cite{SENet,GENet} have been proposed to enhance image classification performance. 
CutMix is complementary to the above methods because it operates on the data level, without changing internal representations or architecture.

\section{CutMix}
\label{sec:cutmix}
\noindent
We describe the CutMix algorithm in detail. 

\subsection{Algorithm}
\label{sec:cutmix_algorithm}
Let $x \in \mathbb{R}^{W \times H \times C}$ and $y$ denote a training image and its label, respectively.
The goal of CutMix is to generate a new training sample $(\Tilde{x},\Tilde{y})$ by combining two training samples $(x_{A}, y_{A})$ and $(x_{B}, y_{B})$. 
{The generated} training sample $(\Tilde{x},\Tilde{y})$ is used to train the model {with} its original loss function.
{We define the combining operation as}
\begin{equation}
\begin{split}
    \Tilde{x} & =  \mathbf{M} \odot x_{A} + (\mathbf{1}- \mathbf{M}) \odot x_{B} \\
    \Tilde{y} & =  \lambda y_A + (1-\lambda) y_B, \\
\end{split}
\label{eq:cutmix}
\end{equation}
where $\mathbf{M} \in \{0,1\}^{W \times H}$ denotes a binary mask indicating where to drop out and fill in from two images, $\mathbf{1}$ is a binary mask filled with ones, and $\odot$ is element-wise multiplication.
Like Mixup~\cite{zhang2017mixup}, the combination ratio $\lambda$ between two data points is sampled from the beta distribution $\text{Beta}(\alpha,\alpha)$. 
In our all experiments, we set $\alpha$ to $1$, that is $\lambda$ is sampled from the uniform distribution $(0,1)$. 
Note that the major difference is that CutMix replaces an image region with a patch from another training image and generates {more locally natural image than Mixup does}.

To sample the binary mask $\mathbf{M}$, we first sample the bounding box coordinates $\mathbf{B}=(r_x,r_y,r_w,r_h)$ indicating the cropping regions on $x_A$ and $x_B$. 
The region $\mathbf{B}$ in $x_A$ is removed and filled in with the patch cropped from $\mathbf{B}$ of $x_B$.

In our experiments, we sample rectangular masks $\mathbf{M}$ whose aspect ratio is proportional to the original image. The box coordinates are uniformly sampled according to:  
\begin{equation}
\begin{split}
    r_{x} \sim \text{Unif}~ \left(0, W\right), &~~~r_{w} = W \sqrt{1-\lambda},  \\
    r_{y} \sim \text{Unif}~ \left(0, H\right), &~~~r_{h} = H \sqrt{1-\lambda}  \\
\end{split}
\label{eq:sampling}
\end{equation}
making the cropped area ratio $\frac{r_w r_h}{WH} = 1-\lambda$.
With the cropping region, the binary mask $\mathbf{M}$ $\in \{0,1\}^{W \times H}$ is decided by filling with $0$ within the bounding box $\mathbf{B}$, otherwise $1$.

In each training iteration, a CutMix-ed sample $(\Tilde{x}, \Tilde{y})$ is generated by combining randomly selected two training samples in a mini-batch according to Equation~(\ref{eq:cutmix}).
Code-level details are presented in Appendix~\ref{appendix:algorithm}. 
CutMix is simple and incurs a negligible computational overhead {as} existing data augmentation techniques used in~\cite{InceptionResnet,densenet}; {we can efficiently utilize it to train any network architecture.}

\subsection{Discussion}
\label{sec:cutmix_discussion}

\begin{figure}
    \centering
    \includegraphics[width=0.77\linewidth]{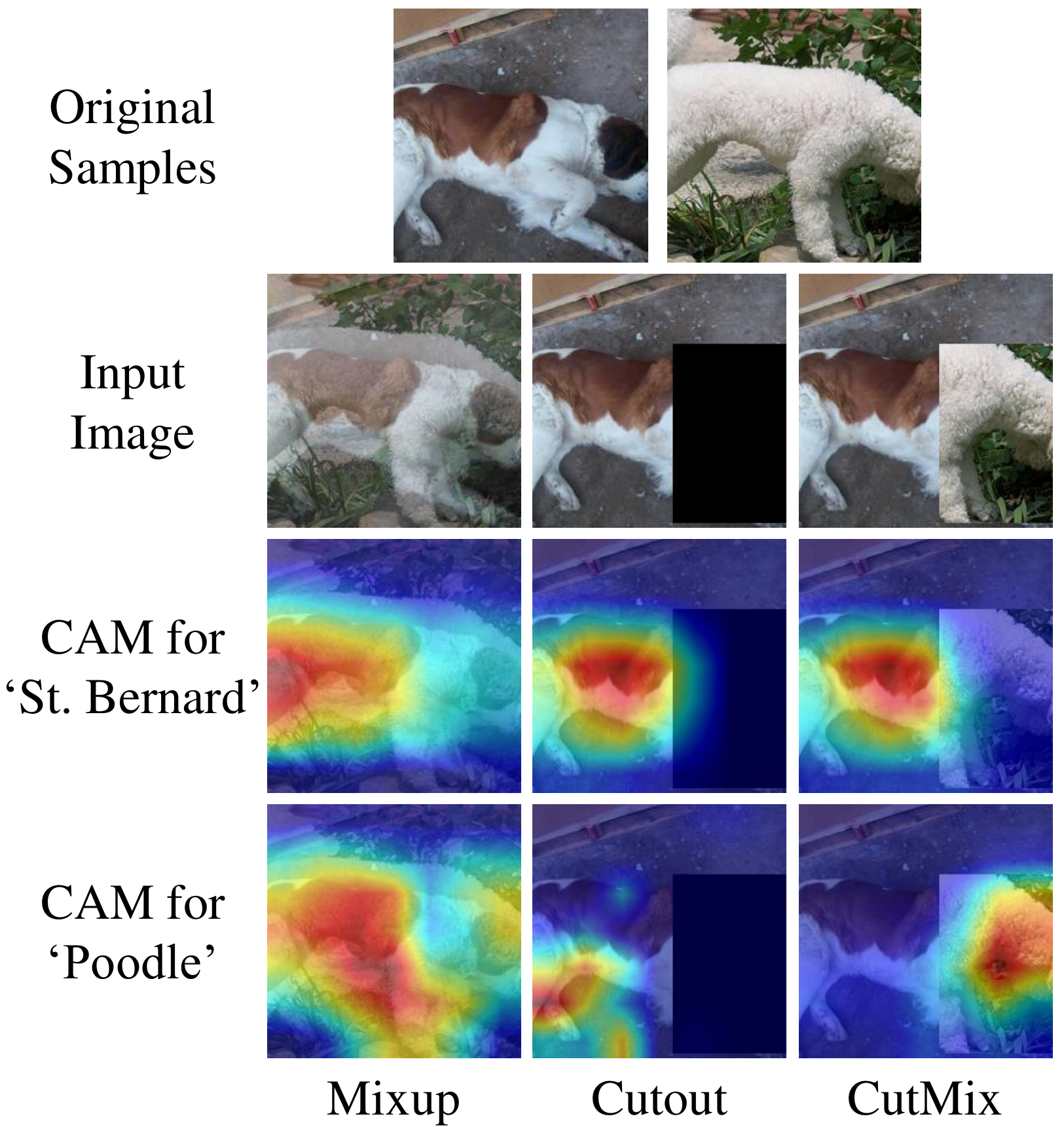}
    \caption{
    Class activation mapping (CAM)~\cite{zhou2016CAM} visualizations on `Saint Bernard' and `Miniature Poodle' samples using various augmentation techniques.
    From top to bottom rows, we show the original images, input augmented image, CAM for class `Saint Bernard', and CAM for class `Miniature Poodle', respectively. 
    {Note that CutMix can take advantage of the mixed region on image, but Cutout cannot.}
    }
    \label{fig:class_cam}
\end{figure}

\noindent\textbf{What does model learn with CutMix? }
We have motivated CutMix such that full object extents are considered as cues for classification, the motivation shared by Cutout, while ensuring two objects are recognized from partial views in a single image to increase training efficiency.
To verify that CutMix is indeed learning to recognize two objects from their respective partial views, we visually compare the activation maps for CutMix against Cutout~\cite{devries2017cutout} and Mixup~\cite{zhang2017mixup}. 
Figure~\ref{fig:class_cam} shows example augmentation inputs as well as corresponding class activation maps (CAM)~\cite{zhou2016CAM} for two classes present, Saint Bernard and Miniature Poodle. 
We {use} vanilla ResNet-50 model\footnote{We use ImageNet-pretrained ResNet-50 provided by PyTorch~\cite{paszke2017automatic}.} for obtaining the CAMs to clearly see the effect of augmentation method only.

\begin{table}[t]
\centering
\tabcolsep=0.1cm
\begin{tabular}{@{}lccc@{}}
\toprule
                            & Mixup & Cutout & CutMix \\ \midrule
Usage of full image region  & \textcolor{darkergreen}{\ding{52}}   & \textcolor{red2}{\ding{56}}      & \textcolor{darkergreen}{\ding{52}}      \\
{Regional dropout} & \textcolor{red2}{\ding{56}}     & \textcolor{darkergreen}{\ding{52}}      & \textcolor{darkergreen}{\ding{52}}    \\
Mixed image \& label & \textcolor{darkergreen}{\ding{52}}    & \textcolor{red2}{\ding{56}}     & \textcolor{darkergreen}{\ding{52}}    \\ 
\bottomrule
\end{tabular}
\caption{Comparison among Mixup, Cutout, and CutMix.}
\label{tab:checkbox}
\end{table}

We observe that Cutout successfully lets a model focus on less discriminative parts of the object, such as the belly of Saint Bernard, while being inefficient due to unused pixels.
Mixup, on the other hand, makes full use of pixels, but introduces unnatural artifacts.
The CAM for Mixup, as a result, shows that the model is confused when choosing cues for recognition.
We hypothesize that such confusion leads to its suboptimal performance in classification and localization, as we will see in {Section~\ref{sec:experiments}.}

{CutMix efficiently improves upon Cutout by being able to localize the two object classes accurately. }
We summarize the key differences among Mixup, Cutout, and CutMix in Table~\ref{tab:checkbox}.

\begin{figure}
    \centering
    \includegraphics[width=\linewidth]{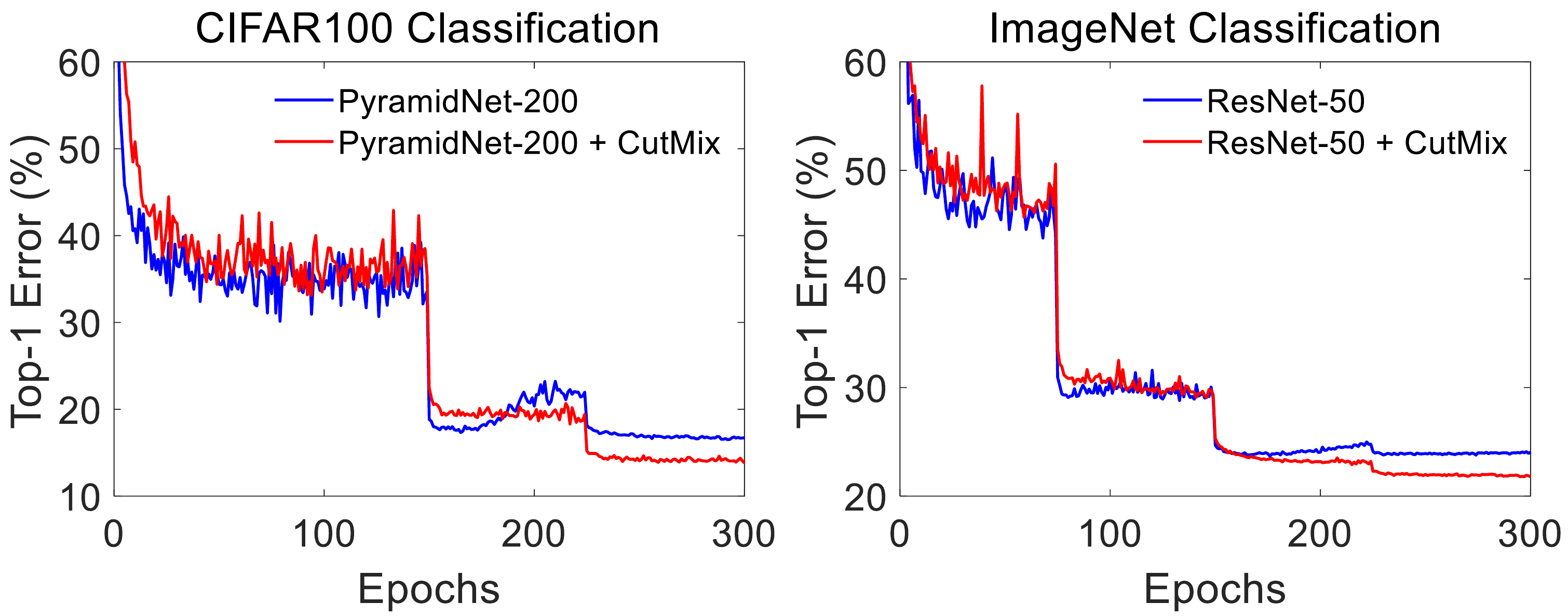}
    \caption{Top-1 test error plot for CIFAR100 (left) and ImageNet (right) classification. 
    Cutmix achieves lower test errors than the baseline at the end of training.
    }
    \label{fig:cls_loss}
\end{figure}

\noindent\textbf{Analysis on validation error: }
We analyze the effect of CutMix on stabilizing the training of deep networks.
We compare the top-1 validation error during the training with CutMix against the baseline.
We train ResNet-50~\cite{resnet} for ImageNet Classification, and PyramidNet-200~\cite{pyramidnet} for CIFAR-100 Classification.
Figure~\ref{fig:cls_loss} shows the results.

We observe, first of all, that CutMix achieves lower validation errors than the baseline at the end of training. 
At epoch 150 when the learning rates are reduced, the baselines suffer from overfitting with increasing validation error.
CutMix, on the other hand, shows a steady decrease in validation error; diverse training samples reduce overfitting.

\section{Experiments}
\label{sec:experiments}
In this section, 
we evaluate CutMix for its {capability} to improve localizability as well as generalizability of {a} trained {model on multiple tasks.}
We first study the effect of CutMix on image classification (Section~\ref{sec:image_cls}) and weakly supervised object localization (Section~\ref{sec:wsol}).
{Next, we show the transferability of a CutMix pre-trained model} when it is fine-tuned for object detection and image captioning tasks (Section~\ref{sec:transfer}). 
{We also show that CutMix can improve the model robustness and alleviate the model over-confidence in Section~\ref{sec:robustness}.}

All experiments were implemented and evaluated on NAVER Smart Machine Learning (NSML)~\cite{nsml} platform with PyTorch~\cite{paszke2017automatic}. Source code and pretrained models are available at \href{https://github.com/clovaai/CutMix-PyTorch}{https://github.com/clovaai/CutMix-PyTorch}.

\subsection{Image Classification}
\label{sec:image_cls}

\subsubsection{ImageNet Classification}
\label{sec:imagenet_cls}
We evaluate on ImageNet-1K benchmark~\cite{ImageNet}, the dataset containing 1.2M training images and 50K validation images of 1K categories.
For fair comparison, we use the standard augmentation setting for ImageNet dataset such as re-sizing, cropping, and flipping, as done in~\cite{pyramidnet,ghiasi2018dropblock,densenet,GoogleNet}. 
We found that regularization methods {including} Stochastic Depth~\cite{stochasticdepth},  Cutout~\cite{devries2017cutout}, Mixup~\cite{zhang2017mixup}, and CutMix require a greater number of training epochs till convergence. 
Therefore, we have trained all the models for $300$ epochs with initial learning rate $0.1$ decayed by factor $0.1$ at epochs $75$, $150$, and $225$. 
The batch size is set to $256$. 
The hyper-parameter $\alpha$ is set to $1$. 
We report the best performances of CutMix and other baselines during training.

\begin{table}[]
\small
\centering
\tabcolsep=0.1cm
\begin{tabular}{@{}lccc@{}}
\toprule
\multirow{2}{*}{Model}            & \multirow{2}{*}{\# Params} & \multirow{2}{*}{\begin{tabular}[c]{@{}c@{}}Top-1 \\ Err (\%)\end{tabular}} & \multirow{2}{*}{\begin{tabular}[c]{@{}c@{}}Top-5 \\ Err (\%)\end{tabular}} \\
                                  &                            &                        &        \\ \midrule
ResNet-152*                                     & 60.3 M     & 21.69    & 5.94\\
ResNet-101 + SE Layer*~\cite{SENet}             & 49.4 M     & 20.94    & 5.50\\
ResNet-101 + GE Layer*~\cite{GENet}             & 58.4 M     & 20.74    & 5.29\\
ResNet-50 + SE Layer*~\cite{SENet}              & 28.1 M     & 22.12    & 5.99\\
ResNet-50 + GE Layer*~\cite{GENet}              & 33.7 M     & 21.88    & 5.80\\ \midrule
ResNet-50 (Baseline)                            & 25.6 M     & 23.68    & 7.05\\
ResNet-50 + Cutout~\cite{devries2017cutout}     & 25.6 M     & 22.93    & 6.66\\
ResNet-50 + StochDepth~\cite{stochasticdepth}   & 25.6 M     & 22.46    & 6.27       \\
ResNet-50 + Mixup~\cite{zhang2017mixup}         & 25.6 M     & 22.58    & 6.40\\
ResNet-50 + Manifold Mixup~\cite{verma2018manifoldmixup}& 25.6 M     & 22.50     & 6.21\\
ResNet-50 + DropBlock*~\cite{ghiasi2018dropblock}   & 25.6 M    & 21.87     & {5.98}\\
ResNet-50 + Feature CutMix                      & 25.6 M    & 21.80  & 6.06\\ 
ResNet-50 + CutMix                              & 25.6 M    & \textbf{{21.40}}    & \textbf{{5.92}}\\  \midrule
\end{tabular}
\vspace{-0.1cm}
\caption{ImageNet classification results based on ResNet-50 model.
`*' denotes results reported in the original papers.
}
\vspace{-0.2cm}
\label{table:imagenet-res50}
\end{table}

We briefly describe the settings for baseline augmentation schemes.
We set the dropping rate of residual blocks to $0.25$ for the best performance of Stochastic Depth~\cite{stochasticdepth}.
The mask size for Cutout~\cite{devries2017cutout} is set to $112\times112$ and the location for dropping out is uniformly sampled. 
The performance of DropBlock~\cite{ghiasi2018dropblock} is from the original paper and the difference from our setting is the training epochs which is set to $270$.
Manifold Mixup~\cite{verma2018manifoldmixup} applies Mixup operation on the randomly chosen internal feature map.
We have tried $\alpha=0.5$ and $1.0$ for Mixup and Manifold Mixup and have chosen $1.0$ which has shown better performances.
It is also possible to extend CutMix to feature-level augmentation (Feature CutMix).
Feature CutMix applies CutMix at a randomly chosen layer per minibatch as Manifold Mixup does.

\begin{table}[]
\small
\centering
\begin{tabular}{@{}lccc@{}}
\toprule
\multirow{2}{*}{Model}     & \multirow{2}{*}{\# Params} & \multirow{2}{*}{\begin{tabular}[c]{@{}c@{}}Top-1 \\ Err (\%)\end{tabular}} & \multirow{2}{*}{\begin{tabular}[c]{@{}c@{}}Top-5 \\ Err (\%)\end{tabular}} \\
                          &                            &                               &  \\ \midrule
ResNet-101 (Baseline)~\cite{resnet}    & 44.6 M    & 21.87     & 6.29\\
ResNet-101 + Cutout~\cite{devries2017cutout} & 44.6 M    & 20.72    & 5.51\\ 
ResNet-101 + Mixup~\cite{zhang2017mixup}    & 44.6 M    & 20.52    & 5.28\\ 
ResNet-101 + CutMix                     & 44.6 M    & \textbf{20.17}    & \textbf{5.24}\\ \midrule
ResNeXt-101 (Baseline)~\cite{resnext}   & 44.1 M    &    21.18   &  5.57 \\
ResNeXt-101 + CutMix                    &  44.1 M    &  \textbf{19.47}  & \textbf{5.03}  \\ 
\toprule
\end{tabular}
\vspace{-0.2cm}
\caption{Impact of CutMix on ImageNet classification for ResNet-101 and ResNext-101.}
\vspace{-0.4cm}
\label{table:imagenet-deeper}
\end{table}

\noindent\textbf{Comparison against baseline augmentations: }
Results are given in Table~\ref{table:imagenet-res50}.
We observe that CutMix achieves the best result, {$\textbf{21.40\%}$} top-1 error, among the considered augmentation strategies.
CutMix outperforms Cutout and Mixup, the two closest approaches to ours, by $+1.53\%$ and $+1.18\%$, respectively.
On the feature level as well, we find CutMix preferable to Mixup, with top-1 errors $21.78\%$ and $22.50\%$, respectively.

\noindent\textbf{Comparison against architectural improvements: }
We have also compared improvements due to CutMix versus architectural improvements (\eg greater depth or additional modules). 
We observe that CutMix improves the performance by $\textbf{+2.28\%}$ while increased depth (ResNet-50 $\rightarrow$ ResNet-152) boosts $+1.99\%$ and SE~\cite{SENet} and GE~\cite{GENet} boosts $+1.56\%$ and $+1.80\%$, respectively.
Note that unlike above architectural boosts improvements due to CutMix come at little or memory or computational time.

\noindent\textbf{CutMix for Deeper Models: }
We have explored the performance of CutMix for the deeper networks, ResNet-101~\cite{resnet} and ResNeXt-101 (32$\times$4d)~\cite{resnext}, on ImageNet.
As seen in Table~\ref{table:imagenet-deeper}, we observe $ \textbf{+1.60\%}$ and $\textbf{+1.71\%}$ respective improvements in top-1 errors due to CutMix.

\subsubsection{CIFAR Classification}
\label{sec:CIFAR-100-classification}
We set mini-batch size to $64$ and training epochs to $300$.
The learning rate was initially set to $0.25$ and decayed by the factor of $0.1$ at $150$ and $225$ epoch.
To ensure the effectiveness of the proposed method, we used a strong baseline, PyramidNet-200~\cite{pyramidnet} with widening factor $\Tilde{\alpha}=240$. It has  $26.8$M parameters and achieves the state-of-the-art performance $16.45\%$ top-1 error on CIFAR-100.

\begin{table}[]
\small
\centering
\begin{tabular}{@{}lcc@{}}
\toprule
\multirow{2}{*}{\begin{tabular}[c]{@{}c@{}}PyramidNet-200 ($\Tilde{\alpha}$=240) \\ (\# params: 26.8 M)\end{tabular}} & \multirow{2}{*}{\begin{tabular}[c]{@{}c@{}}Top-1 \\ Err (\%)\end{tabular}} & \multirow{2}{*}{\begin{tabular}[c]{@{}c@{}}Top-5 \\ Err (\%)\end{tabular}} \\  & & \\ \midrule
Baseline                                      & 16.45   & 3.69                  \\
+ StochDepth~\cite{stochasticdepth}                                  & 15.86   & 3.33\\
+ Label smoothing ($\epsilon$=0.1)~\cite{szegedy2016rethinking_labelsm} & 16.73   & 3.37\\
+ Cutout~\cite{devries2017cutout}                                      & 16.53   & 3.65\\
+ Cutout + Label smoothing ($\epsilon$=0.1)                    & 15.61   & 3.88\\
+ DropBlock~\cite{ghiasi2018dropblock}                          & 15.73   & 3.26\\
+ DropBlock + Label smoothing ($\epsilon$=0.1)                  & 15.16   & 3.86\\
+ Mixup ($\alpha$=0.5)~\cite{zhang2017mixup}                    & 15.78   & 4.04\\
+ Mixup ($\alpha$=1.0)~\cite{zhang2017mixup}                    & 15.63   & 3.99\\
+ Manifold Mixup ($\alpha$=1.0)~\cite{verma2018manifoldmixup}    & 16.14   & 4.07\\
+ Cutout + Mixup ($\alpha$=1.0)                & 15.46   & 3.42                  \\
+ Cutout + Manifold Mixup ($\alpha$=1.0)        & 15.09   & 3.35                  \\
+ ShakeDrop~\cite{yamada2018shakedrop}         & 15.08   & {2.72}         \\
+ CutMix                           & {14.47}                      & 2.97\\
+ CutMix  + ShakeDrop~\cite{yamada2018shakedrop} & \textbf{{13.81}}          & \textbf{2.29} \\  \toprule
\end{tabular}
\vspace{-0.2cm}
\caption{Comparison of state-of-the-art regularization methods on CIFAR-100. 
}
\vspace{-0.4cm}
\label{table:cifar-pyramidnet}
\end{table}

Table~\ref{table:cifar-pyramidnet} shows the performance comparison against other state-of-the-art data augmentation and regularization methods. 
All experiments were conducted three times and the averaged best performances during training are reported.

\noindent\textbf{Hyper-parameter settings:}
We set the hole size of Cutout~\cite{devries2017cutout} to $16\times16$.
For DropBlock~\cite{ghiasi2018dropblock}, \texttt{keep\_prob} and \texttt{block\_size} are set to $0.9$ and $4$, respectively.
The drop rate for Stochastic Depth~\cite{stochasticdepth} is set to 0.25. 
For Mixup~\cite{zhang2017mixup}, we tested the hyper-parameter $\alpha$ with 0.5 and 1.0.
For Manifold Mixup~\cite{verma2018manifoldmixup}, we applied Mixup operation at a randomly chosen layer per minibatch.

\begin{table}[]
\small
\centering
\begin{tabular}{@{}lccc@{}}
\toprule
\multirow{2}{*}{Model}  & \multirow{2}{*}{\# Params} & \multirow{2}{*}{\begin{tabular}[c]{@{}c@{}}Top-1 \\ Err (\%)\end{tabular}} & \multirow{2}{*}{\begin{tabular}[c]{@{}c@{}}Top-5 \\ Err (\%)\end{tabular}} \\
                        &                       &                          & \\ \midrule
PyramidNet-110 ($\Tilde{\alpha}=64$)~\cite{pyramidnet}    &    1.7 M & 19.85  & 4.66    \\
PyramidNet-110 + CutMix &                1.7 M     &  \textbf{17.97}      &     \textbf{3.83}    \\ \midrule
ResNet-110~\cite{resnet}              &     1.1 M           & 23.14          & 5.95    \\
ResNet-110 + CutMix     &     1.1 M            & \textbf{20.11}          & \textbf{4.43}    \\ 
\toprule
\end{tabular}
\vspace{-0.2cm}
\caption{Impact of CutMix on lighter architectures on CIFAR-100.}
\vspace{-0.3cm}
\label{table:cifar-various}
\end{table}

\begin{table}[]
\centering
\begin{tabular}{@{}lcc@{}}
\toprule
PyramidNet-200 ($\Tilde{\alpha}$=240)  & Top-1 Error (\%) \\ 
\midrule
Baseline                                                                & 3.85  \\
+ Cutout                                                            & 3.10   \\
+ Mixup ($\alpha$=1.0)                                                 & 3.09  \\
+ Manifold Mixup ($\alpha$=1.0)                                     &  3.15 \\
+ CutMix                                                                & \textbf{2.88} \\  \toprule
\end{tabular}
\vspace{-0.1cm}
\caption{Impact of CutMix on CIFAR-10. 
}
\label{table:cifar-10}
\vspace{-0.3cm}
\end{table}

\noindent\textbf{Combination of regularization methods:}
We have evaluated the combination of regularization methods.
Both Cutout~\cite{devries2017cutout} and label smoothing~\cite{szegedy2016rethinking_labelsm} does not improve the accuracy when adopted independently, but they are effective when used together.
Dropblock~\cite{ghiasi2018dropblock}, the feature-level generalization of Cutout, is also more effective when label smoothing is also used.
Mixup~\cite{zhang2017mixup} and Manifold Mixup~\cite{verma2018manifoldmixup} achieve higher accuracies when Cutout is applied on input images.
{The combination of Cutout and Mixup tends to generate locally separated and mixed samples since the cropped regions have less ambiguity than those of the vanilla Mixup. The superior performance of Cutout and Mixup combination shows that mixing via cut-and-paste manner is better than interpolation, as much evidenced by CutMix performances.
}

{CutMix achieves $14.47\%$ top-1 classification error on CIFAR-100, $+1.98\%$ higher than the baseline performance $16.45\%$.}
{We have achieved a new state-of-the-art performance $13.81\%$ by combining CutMix and ShakeDrop~\cite{yamada2018shakedrop}, a regularization that adds noise on intermediate features.}

\noindent\textbf{CutMix for various models: }
Table~\ref{table:cifar-various} shows CutMix also significantly improves the performance of the weaker baseline architectures, such as PyramidNet-110~\cite{pyramidnet} and ResNet-110.

\noindent\textbf{CutMix for CIFAR-10: }
We have evaluated CutMix on CIFAR-10 dataset using the same baseline and training setting for CIFAR-100.
The results are given in Table~\ref{table:cifar-10}.
On CIFAR-10, CutMix also enhances the classification performances by $+0.97\%$, outperforming Mixup and Cutout performances.

\subsubsection{Ablation Studies}
\label{sec:ablation}
\begin{figure}[t]
    \centering
    \includegraphics[width=\linewidth]{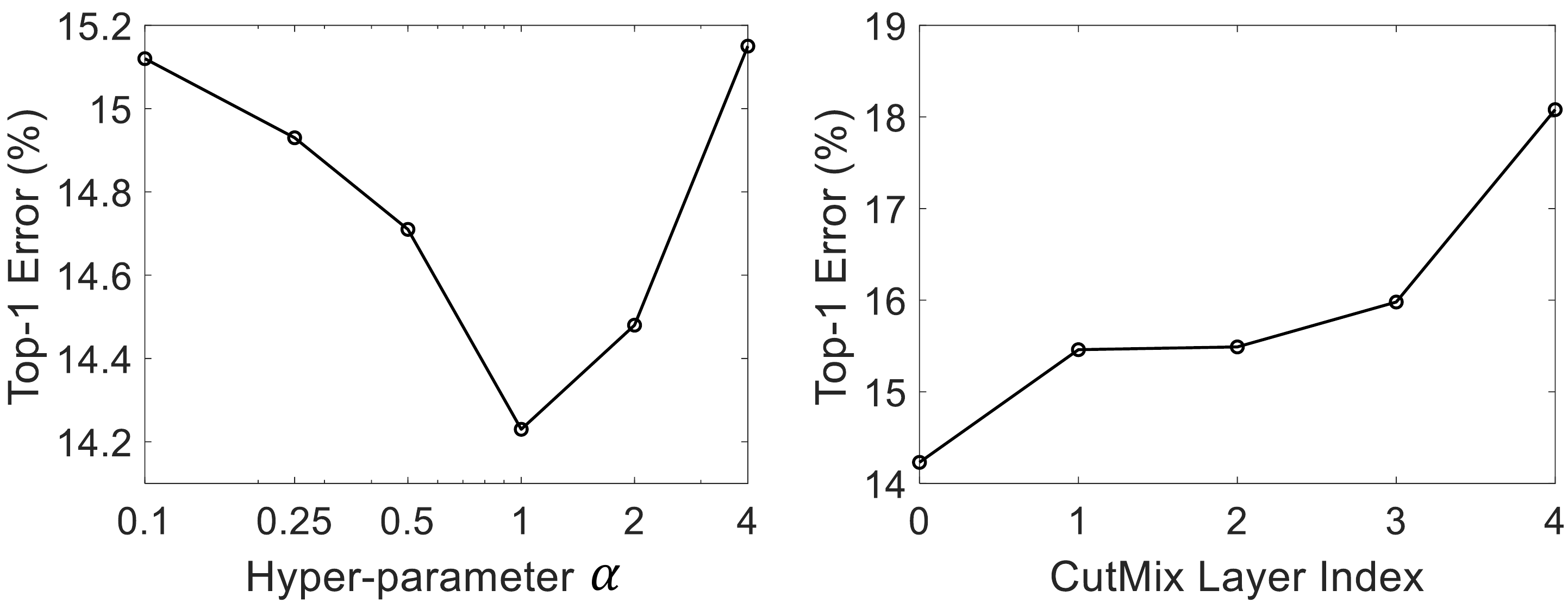}
    \caption{Impact of $\alpha$ and CutMix layer depth on CIFAR-100 top-1 error.
    }
    \label{fig:ablation}
\end{figure}

We conducted ablation study in CIFAR-100 dataset using the same experimental settings in Section \ref{sec:CIFAR-100-classification}.
We evaluated CutMix with $\alpha\in\{0.1,0.25,0.5,1.0,2.0,4.0\}$; the results are given in Figure~\ref{fig:ablation}, left plot.
For all $\alpha$ values considered, CutMix improves upon the baseline ($16.45\%$). The best performance is achieved when $\alpha=1.0$.

The performance of feature-level CutMix is given in Figure~\ref{fig:ablation}, right plot.
We changed the layer on which CutMix is applied, from image layer itself to higher feature levels. 
We denote the index as (0=image level, 1=after first \texttt{conv-bn}, 2=after \texttt{layer1}, 3=after \texttt{layer2}, 4=after \texttt{layer3}).
CutMix achieves the best performance when it is applied on the input images. 
Again, feature-level CutMix except the \texttt{layer3} case improves the accuracy over the baseline ($16.45\%$).

\begin{table}[]
\centering
\begin{tabular}{@{}lcc@{}}
\toprule
\multirow{2}{*}{\begin{tabular}[c]{@{}c@{}}PyramidNet-200 ($\Tilde{\alpha}$=240) \\ ($\#$ params: 26.8 M)\end{tabular}} & \multirow{2}{*}{\begin{tabular}[c]{@{}c@{}}Top-1 \\ Error (\%)\end{tabular}} & \multirow{2}{*}{\begin{tabular}[c]{@{}c@{}}Top-5 \\ Error (\%)\end{tabular}} \\  & & \\ \midrule
Baseline                                      & 16.45               & 3.69         \\
Proposed (CutMix)                                    & 14.47               & 2.97         \\ \midrule
Center Gaussian CutMix                        & 15.95               & 3.40         \\
Fixed-size CutMix                           &   14.97               &  3.15 \\ 
One-hot CutMix                              &  15.89             &  3.32             \\
Scheduled CutMix                             & 14.72               & 3.17         \\
Complete-label CutMix                           & 15.17               & 3.10         \\
\toprule                                                     
\end{tabular}
\vspace{-0.2cm}
\caption{Performance of CutMix variants on CIFAR-100.}
\vspace{-0.4cm}
\label{table:ablation}
\end{table}

We explore different design choices for CutMix.
Table~\ref{table:ablation} shows the performance of CutMix variations.
`Center Gaussian CutMix' samples the box coordinates $r_x,r_y$ of Equation~(\ref{eq:sampling}) according to the Gaussian distribution with mean at the image center, instead of the original uniform distribution.
`Fixed-size CutMix' fixes the size of cropping region $(r_w,r_h)$ at $16 \times 16$ (i.e. $\lambda=0.75$).
`Scheduled CutMix' linearly increases the probability to apply CutMix as training progresses, as done by \cite{ghiasi2018dropblock,stochasticdepth}, from $0$ to $1$.
`One-hot CutMix' decides the mixed target label by committing to the label of greater patch portion (single one-hot label), rather than using the combination strategy in Equation~(\ref{eq:cutmix}).
`Complete-label CutMix' assigns the mixed target label as $\Tilde{y} =  0.5y_A + 0.5y_B$ regardless of the combination ratio $\lambda$.
The results show that above variations lead to performance degradation compared to the original CutMix.

\subsection{Weakly Supervised Object Localization}
\label{sec:wsol}

\begin{table}[]
\centering
\tabcolsep=0.1cm
\begin{tabular}{@{}lcc@{}}
\toprule
Method             & \begin{tabular}[c]{@{}c@{}}CUB200-2011\\ Loc Acc (\%) \end{tabular} & \begin{tabular}[c]{@{}c@{}}ImageNet\\ Loc Acc (\%) \end{tabular} \\ 
\midrule
VGG-GAP + CAM~\cite{zhou2016CAM} & 37.12  & 42.73            \\
VGG-GAP + ACoL*~\cite{zhang2018ACOL} & 45.92  & {45.83}    \\
VGG-GAP + ADL*~\cite{choe2019attention}    & 52.36  & 44.92    \\
GoogLeNet + HaS*~\cite{singh2017hide} &  -  & 45.21            \\
InceptionV3 + SPG*~\cite{zhang2018SPG} & 46.64 & 48.60\\
\midrule
VGG-GAP + Mixup~\cite{zhang2017mixup}     & 41.73    & 42.54    \\
VGG-GAP + Cutout~\cite{devries2017cutout} & 44.83    & 43.13    \\
VGG-GAP + CutMix                    & {\textbf{52.53}}     & \textbf{43.45}    \\
\midrule
ResNet-50 + CAM~\cite{zhou2016CAM} & 49.41  & 46.30            \\
ResNet-50 + Mixup~\cite{zhang2017mixup}     & 49.30    & 45.84      \\
ResNet-50 + Cutout~\cite{devries2017cutout} & 52.78    & 46.69       \\
ResNet-50 + CutMix                  & {\textbf{54.81}}     & \textbf{47.25}    \\ \bottomrule
\end{tabular}
\caption{Weakly supervised object localization results on CUB200-2011 and ImageNet. * denotes results reported in the original papers.
}
\label{table:WSOL} 
\end{table}

\begin{table*}[]
\centering
\begin{tabular}{lccccc}
\hline
\multirow{3}{*}{\begin{tabular}[c]{@{}l@{}}Backbone \\ Network\end{tabular}} 
& \multicolumn{1}{c}{\multirow{3}{*}{\begin{tabular}[c]{@{}c@{}}ImageNet Cls\\ Top-1 Error (\%)\end{tabular}}} 
& \multicolumn{2}{c}{Detection} 
& \multicolumn{2}{c}{Image Captioning} \\ \cline{3-6} 
& \multicolumn{1}{c}{}                                                                                         
& \multirow{2}{*}{\begin{tabular}[c]{@{}c@{}}SSD~\cite{liu2016ssd}\\ (mAP)\end{tabular}} 
& \multirow{2}{*}{\begin{tabular}[c]{@{}c@{}}Faster-RCNN~\cite{fasterrcnn}\\ (mAP)\end{tabular}} 
& \multirow{2}{*}{\begin{tabular}[c]{@{}c@{}}NIC~\cite{vinyals2015show} \\ (BLEU-1)\end{tabular}} 
& \multirow{2}{*}{\begin{tabular}[c]{@{}c@{}}NIC~\cite{vinyals2015show} \\ (BLEU-4)\end{tabular}} \\
& \multicolumn{1}{c}{} 
& & & & \\ \hline
ResNet-50 (Baseline)& 23.68 & 76.7 (+0.0)         & 75.6 (+0.0)     & 61.4 (+0.0)        & 22.9 (+0.0)       \\
Mixup-trained       & 22.58 & 76.6 \redp{-0.1} & 73.9 \redp{-1.7} & 61.6 \greenp{+0.2} & 23.2  \greenp{+0.3} \\
Cutout-trained      & 22.93 & 76.8 \greenp{+0.1} & 75.0 \redp{-0.6} & 63.0 \greenp{+1.6} & 24.0 \greenp{+1.1} \\
CutMix-trained      & 21.40 & \textbf{77.6  \greenp{+0.9}} & \textbf{76.7  \greenp{+1.1}} & \textbf{64.2  \greenp{+2.8}} & \textbf{24.9  \greenp{+2.0}} \\ \hline
\end{tabular}
\vspace{-0.15cm}
\caption{Impact of CutMix on transfer learning of pretrained model to other tasks, object detection and image captioning.}
\label{table:imagenet-detection}
\vspace{-0.4cm}
\end{table*}

Weakly supervised object localization (WSOL) task aims to train the classifier to localize target objects by using only the class labels. 
To localize the target well, it is important to make CNNs extract cues from full object regions and not focus on small discriminant parts of the target.
Learning spatially distributed representation is thus the key for improving performance on WSOL task. 
CutMix guides a classifier to attend to broader sets of cues to make decisions; we expect CutMix to improve WSOL performances of classifiers. 
To measure this, we apply CutMix over baseline WSOL models.
We followed the training and evaluation strategy of existing WSOL methods \cite{zhang2018ACOL, zhang2018SPG,choe2019attention} with
VGG-GAP and ResNet-50 as the base architectures.
The quantitative and qualitative results are given in Table~\ref{table:WSOL} and Figure~\ref{fig:WSOL}, respectively.
Full implementation details are in Appendix~\ref{appendix:wsol}. 

\noindent\textbf{Comparison against Mixup and Cutout: }
CutMix outperforms Mixup~\cite{zhang2017mixup} on localization accuracies by $+5.51\%$ and $+1.41\%$ on CUB200-2011 and ImageNet, respectively. 
Mixup degrades the localization accuracy of the baseline model; it tends to make a classifier focus on small regions as shown in Figure~\ref{fig:WSOL}. 
{As we have hypothesized in Section~\ref{sec:cutmix_discussion}, more ambiguity in Mixup samples make a classifier focus on even more discriminative parts of objects, leading to decreased localization accuracies.}
Although Cutout~\cite{devries2017cutout} improves the accuracy over the baseline, it is outperformed by CutMix: $+2.03\%$ and $+0.56\%$ on CUB200-2011 and ImageNet, respectively. 

CutMix also achieves comparable localization accuracies on CUB200-2011 and ImageNet, even when compared against the dedicated state-of-the-art WSOL methods~\cite{zhou2016CAM,singh2017hide,zhang2018ACOL,zhang2018SPG,choe2019attention} that focus on learning spatially dispersed representations. 

\begin{figure}
    \centering
    \includegraphics[width=\linewidth]{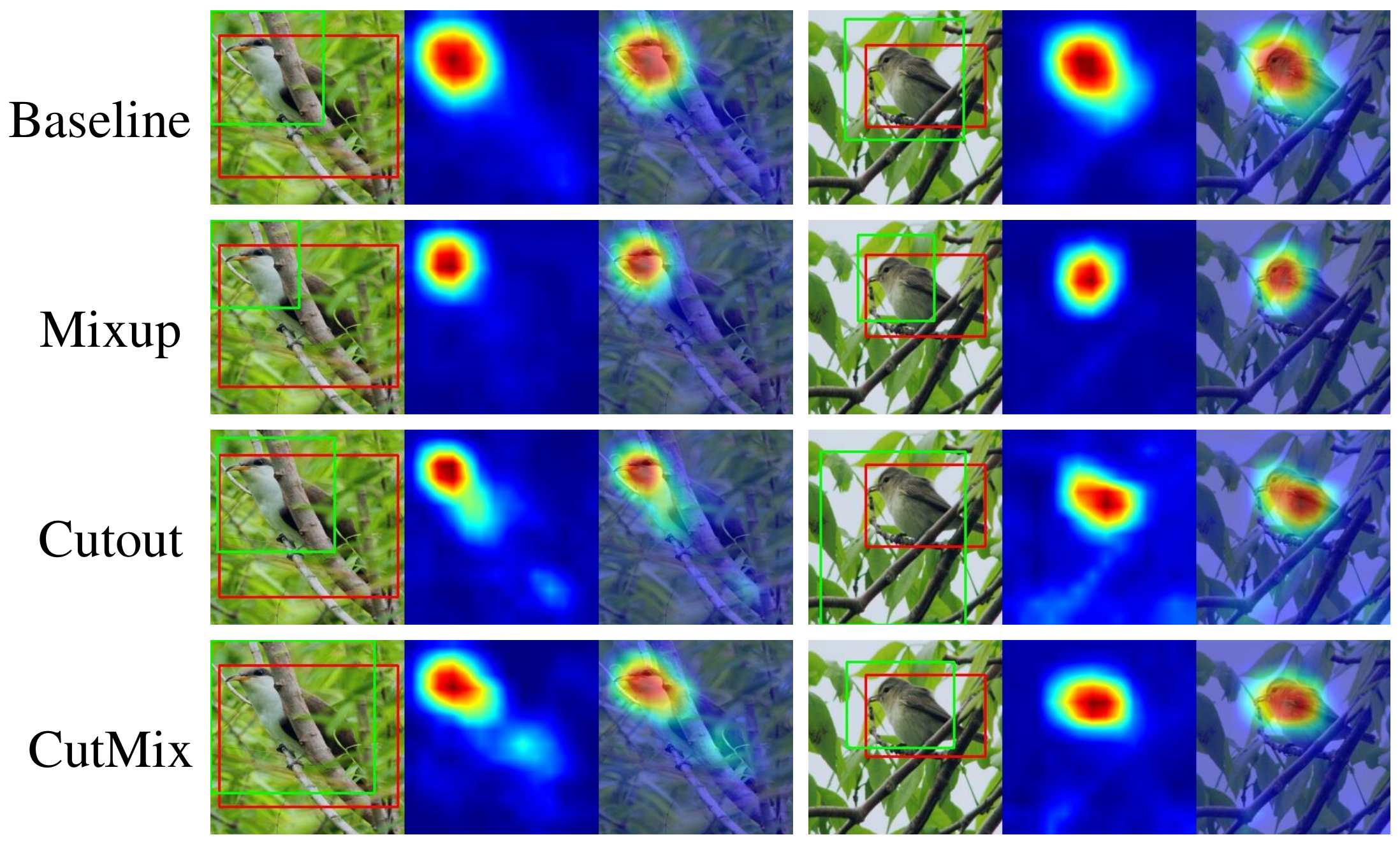}
    \caption{Qualitative comparison of the baseline (ResNet-50), Mixup, Cutout, and CutMix for weakly supervised object localization task on CUB-200-2011 dataset. Ground truth and predicted bounding boxes are denoted as red and green, respectively.
    }
    \label{fig:WSOL}
    \vspace{-0.2cm}
\end{figure}

\subsection{Transfer Learning of Pretrained Model}
\label{sec:transfer}

ImageNet pre-training is de-facto standard practice for many visual recognition tasks.
We examine whether CutMix pre-trained models leads to better performances in certain downstream tasks based on ImageNet pre-trained models.
As CutMix has shown superiority in localizing less discriminative object parts, we would expect it to lead to boosts in certain recognition tasks with localization elements, such as object detection and image captioning.
We evaluate the boost from CutMix on those tasks by replacing the backbone network initialization with other ImageNet pre-trained models using Mixup~\cite{zhang2017mixup}, Cutout~\cite{devries2017cutout}, and CutMix.
ResNet-50 is used as the baseline architecture in this section. 

\noindent\textbf{
Transferring to Pascal VOC object detection: 
}
Two popular detection models, SSD~\cite{liu2016ssd} and Faster RCNN~\cite{fasterrcnn}, are considered.
Originally the two methods have utilized VGG-16 as backbones, but we have changed it to ResNet-50.
The ResNet-50 backbone is initialized with various ImageNet-pretrained models and then fine-tuned on Pascal VOC 2007 and 2012~\cite{pascalvoc} \texttt{trainval} data. Models are evaluated on VOC 2007 \texttt{test} data using the mAP metric.
We follow the fine-tuning strategy of the original methods~\cite{liu2016ssd,fasterrcnn}; implementation details are in Appendix~\ref{appendix:Detection}. 
Results are shown in Table~\ref{table:imagenet-detection}. 
Pre-training with Cutout and Mixup has failed to improve the object detection performance over the vanilla pre-trained model. 
However, the pre-training with CutMix improves the performance of both SSD and Faster-RCNN. 
Stronger localizability of the CutMix pre-trained models leads to better detection performances.

\begin{figure*}[ht!]
    \centering
    \begin{subfigure}[]{0.495\linewidth}
        \begin{subfigure}[ht!]{0.495\linewidth}
            \includegraphics[width=\linewidth]{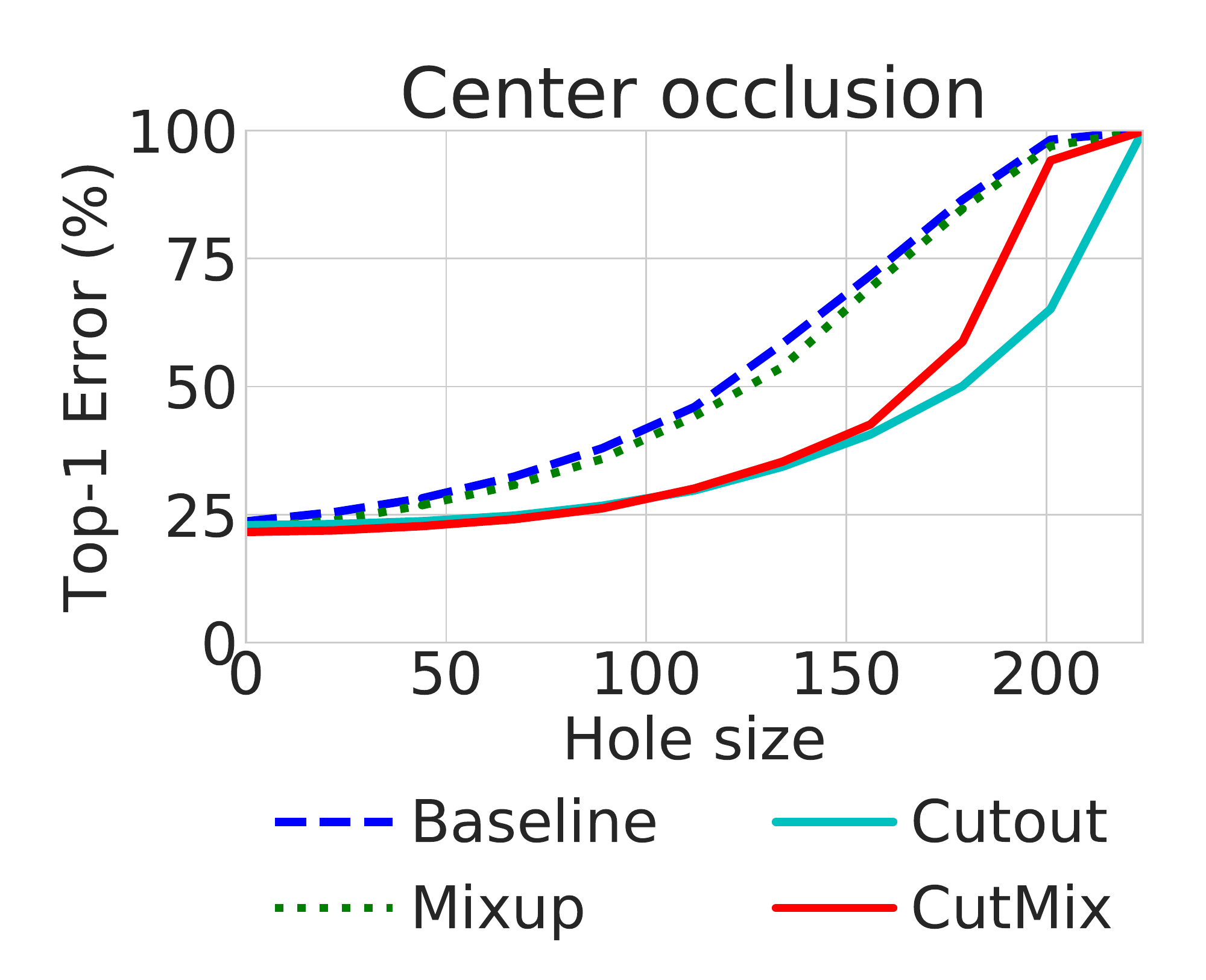}
        \end{subfigure}
        \begin{subfigure}[ht!]{0.495\linewidth}
            \includegraphics[width=\linewidth]{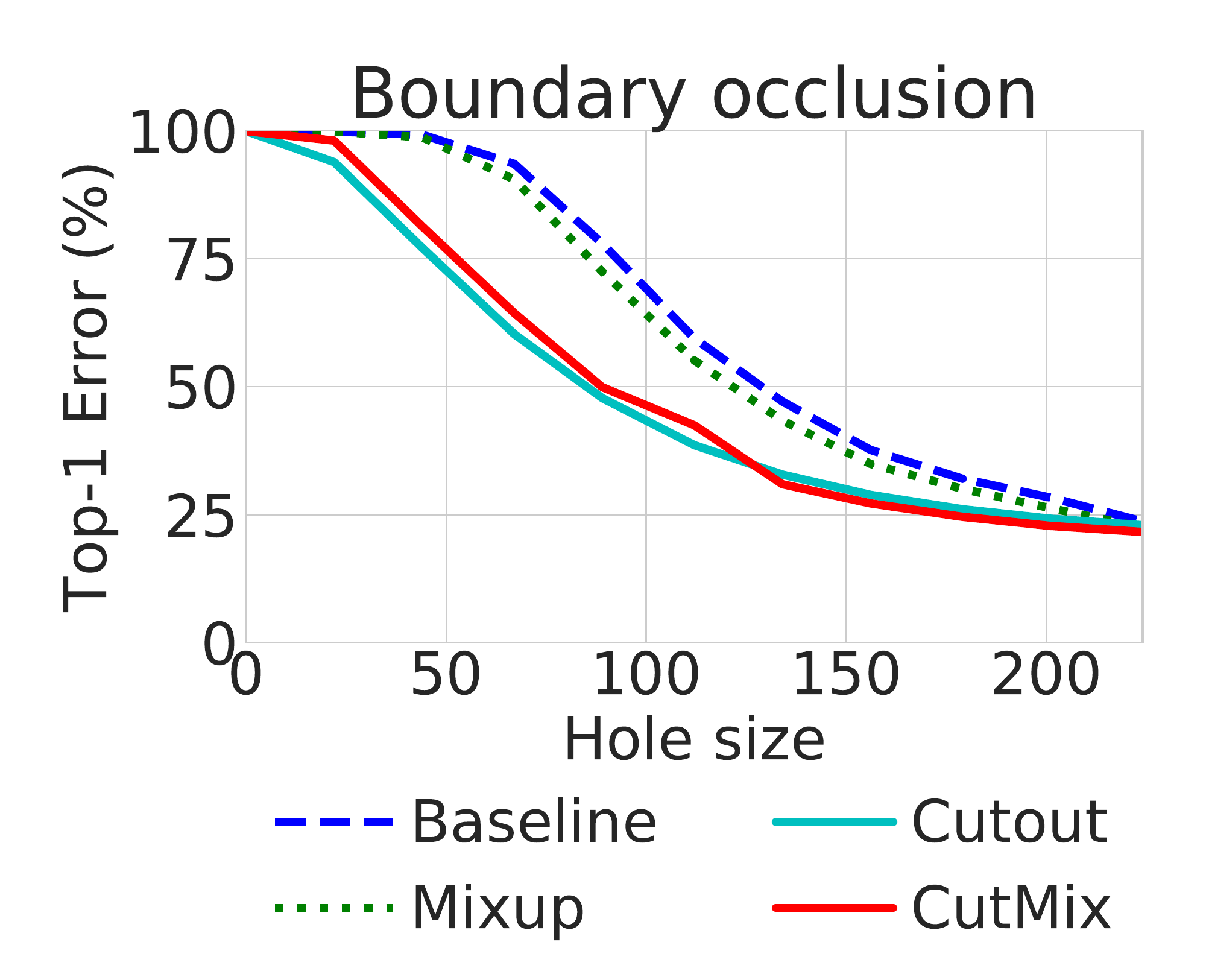}
        \end{subfigure}
        \caption{Analysis for occluded samples}
        \label{fig:robustness_occ}
    \end{subfigure}
    \begin{subfigure}[]{0.495\linewidth}
        \begin{subfigure}[ht!]{0.495\linewidth}
            \includegraphics[width=\linewidth]{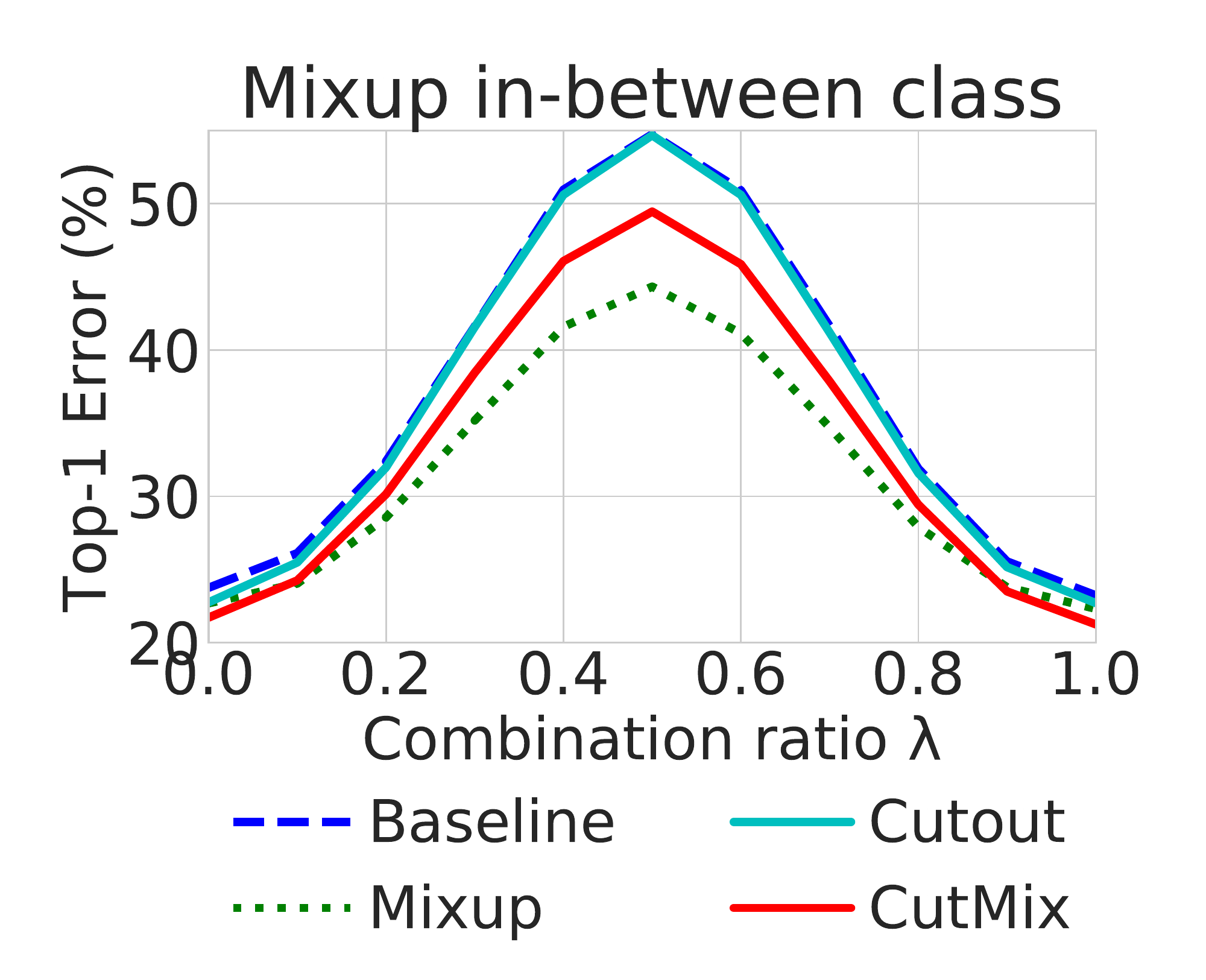}
        \end{subfigure}
        \begin{subfigure}[ht!]{0.495\linewidth}
            \includegraphics[width=\linewidth]{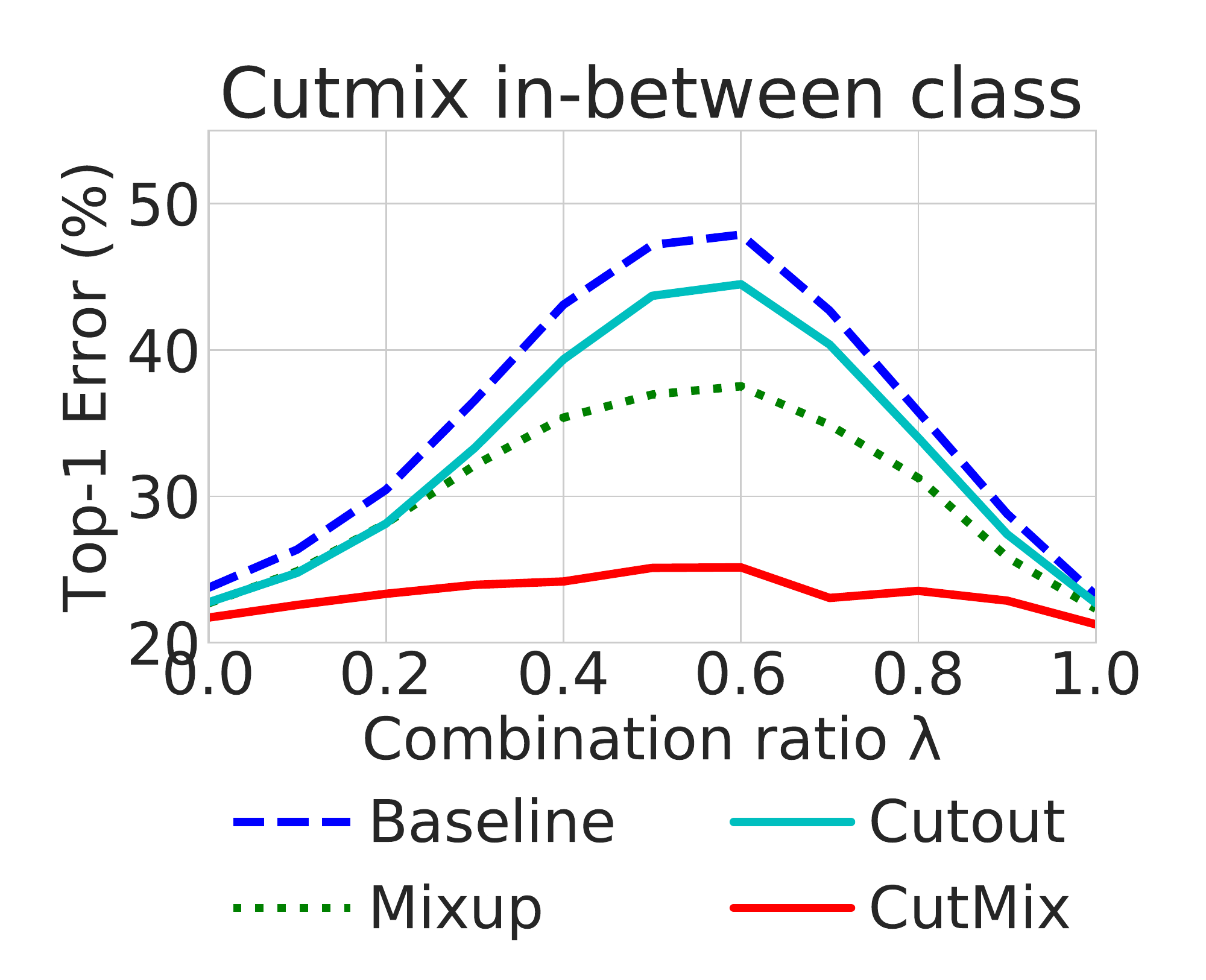}
        \end{subfigure}
        \caption{Analysis for in-between class samples}
        \label{fig:robustness_inbetween}
    \end{subfigure}
    \vspace{-0.1cm}
    \caption{Robustness experiments on the ImageNet validation set.}
    \label{fig:robustness}
    \vspace{-0.3cm}
\end{figure*}

\noindent\textbf{
Transferring to MS-COCO image captioning: 
}
We used Neural Image Caption (NIC)~\cite{vinyals2015show} as the base model for image captioning experiments. We have changed the backbone network of encoder from GoogLeNet~\cite{vinyals2015show} to ResNet-50.
The backbone network is initialized with various ImageNet pre-trained models, and then trained and evaluated on MS-COCO dataset~\cite{lin2014microsoft}.
Implementation details and evaluation metrics (METEOR, CIDER, etc.) are in Appendix~\ref{appendix:captioning}.
Table~\ref{table:imagenet-detection} shows the results.
CutMix outperforms Mixup and Cutout in both \texttt{BLEU1} and  \texttt{BLEU4} metrics. 
Simply replacing backbone network with our CutMix pre-trained model gives performance gains for object detection and image captioning tasks at no extra cost.

\subsection{Robustness and Uncertainty}
\label{sec:robustness}

Many researches have shown that deep models are easily fooled by small and unrecognizable perturbations on the input images, a phenomenon referred to as adversarial attacks~\cite{fgsm, szegedy2013intriguing}.
One straightforward way to enhance robustness and uncertainty is an input augmentation by generating unseen samples~\cite{madry2017towards}.
We evaluate robustness and uncertainty improvements due to input augmentation methods including Mixup, Cutout, and CutMix.

\noindent\textbf{Robustness: }
We evaluate the robustness of the trained models to adversarial samples, occluded samples, and in-between class samples.
We use ImageNet pre-trained ResNet-50 models with same setting as in Section~\ref{sec:imagenet_cls}.

Fast Gradient Sign Method (FGSM)~\cite{fgsm} is used to generate adversarial perturbations and we assume that the adversary has full information of the models (\textit{white-box attack}).
We report top-1 accuracies after attack on ImageNet validation set in Table~\ref{table:fgsm}. 
CutMix significantly improves the robustness to adversarial attacks compared to other augmentation methods.

For occlusion experiments, we generate occluded samples in two ways: center occlusion by filling zeros in a center hole and boundary occlusion by filling zeros outside of the hole.
In Figure~\ref{fig:robustness_occ}, we measure the top-1 error by varying the hole size from $0$ to $224$. For both occlusion scenarios, Cutout and CutMix achieve significant improvements in robustness while Mixup only marginally improves it.
Interestingly, CutMix almost achieves a comparable performance as Cutout even though CutMix has not observed any occluded sample during training unlike Cutout.

Finally, we evaluate the top-1 error of Mixup and CutMix in-between samples. 
The probability to predict neither two classes by varying the combination ratio $\lambda$ is illustrated in Figure~\ref{fig:robustness_inbetween}.
We randomly select $50,000$ in-between samples in ImageNet validation set.
In both experiments, Mixup and CutMix improve the performance while improvements due to Cutout are almost negligible.
Similarly to the previous occlusion experiments, CutMix even improves the robustness to the unseen Mixup in-between class samples.

\begin{table}[]
\centering
\begin{tabular}{@{}lcccc@{}}
\toprule
               & Baseline & Mixup & Cutout & CutMix        \\ \midrule
Top-1 Acc (\%) & 8.2      & 24.4  & 11.5   & \textbf{31.0} \\ \bottomrule
\end{tabular}
\caption{Top-1 accuracy after FGSM white-box attack on ImageNet validation set.}
\label{table:fgsm}
\end{table}

\begin{table}[]
\tabcolsep=0.1cm
\centering
\begin{tabular}{@{}lccc@{}}
\toprule
Method   & TNR at TPR 95\%              & AUROC                       & Detection Acc. \\ \midrule
Baseline & 26.3 (+0)                    & 87.3 (+0)                   & 82.0 (+0)                    \\
Mixup    & 11.8 \redp{-14.5}            & 49.3 \redp{-38.0}           & 60.9 \redp{-21.0}           \\
Cutout   & 18.8 \redp{-7.5}            & 68.7 \redp{-18.6}            & 71.3 \redp{-10.7}            \\
CutMix   & \textbf{69.0 \greenp{+42.7}} & \textbf{94.4 \greenp{+7.1}} & \textbf{89.1 \greenp{+7.1}} \\ \bottomrule
\end{tabular}
\vspace{-0.15cm}
\caption{Out-of-distribution (OOD) detection results with CIFAR-100 trained models. Results are averaged on seven datasets. All numbers are in percents; higher is better. 
}
\vspace{-0.3cm}
\label{table:OOD}
\end{table}

\noindent\textbf{Uncertainty: }
We measure the performance of the out-of-distribution (OOD) detectors proposed by~\cite{hendrycks2016baseline} which determines whether the sample is in- or out-of-distribution by score thresholding.
We use PyramidNet-200 trained on CIFAR-100 datasets with same setting as in Section~\ref{sec:CIFAR-100-classification}.
In Table~\ref{table:OOD}, we report the averaged OOD detection performances against seven out-of-distribution samples from~\cite{hendrycks2016baseline, liang2017odin}, including TinyImageNet, LSUN~\cite{yu2015lsun}, uniform noise, Gaussian noise, etc. More results are illustrated in Appendix~\ref{appendix:robustness}.
Mixup and Cutout augmentations aggravate the over-confidence of the base networks. Meanwhile, CutMix significantly alleviates the over-confidence of the model.

\section{Conclusion}

We have introduced CutMix for training CNNs with strong classification and localization ability.
CutMix is easy to implement and has no computational overhead, while being surprisingly effective on various tasks.
On ImageNet classification, applying CutMix to ResNet-50 and ResNet-101 brings $+2.28\%$ and $+1.70\%$ top-1 accuracy improvements.
On CIFAR classification, CutMix significantly improves the performance of  baseline by $+1.98\%$ leads to the state-of-the-art top-1 error $14.47\%$.
On weakly supervised object localization (WSOL), CutMix substantially enhances the localization accuracy and has achieved comparable localization performances as the state-of-the-art WSOL methods. 
Furthermore, simply using CutMix-ImageNet-pretrained model as the initialized backbone of the object detection and image captioning brings overall performance improvements.
Finally, we have shown that CutMix results in improvements in robustness and uncertainty of image classifiers over the vanilla model as well as other regularized models.

\section*{Acknowledgement}
We would like to thank Clova AI Research team, especially Jung-Woo Ha and Ziad Al-Halah for their helpful feedback and discussion.


{\small
\bibliographystyle{ieee_fullname}
\bibliography{ms}
}

\clearpage
\setcounter{table}{0}
\renewcommand{\thetable}{A\arabic{table}}
\renewcommand{\thefigure}{A\arabic{figure}}
\renewcommand{\thealgorithm}{A\arabic{algorithm}}

\appendix

\section{CutMix Algorithm}
\label{appendix:algorithm}

\algnewcommand\Input{\item[\textbf{Input:}]}%
\algnewcommand\Output{\item[\textbf{Output:}]}%
\begin{algorithm*}[t]
  \caption{Pseudo-code of CutMix}
  \begin{algorithmic}[1]
    \For {each iteration}
    \State input, target = get\_minibatch(dataset) \Comment{input is N$\times$C$\times$W$\times$H size tensor, target is N$\times$K size tensor.}
        \If{mode $==$ training}
            \State input\_s, target\_s = shuffle\_minibatch(input, target) \Comment{CutMix starts here.}
            \State lambda = Unif(0,1)
            \State r\_x = Unif(0,W)
            \State r\_y = Unif(0,H)
            \State r\_w = Sqrt(1 - lambda)
            \State r\_h = Sqrt(1 - lambda)
            \State x1 = Round(Clip(r\_x - r\_w / 2, min=0))
            \State x2 = Round(Clip(r\_x + r\_w / 2, max=W))
            \State y1 = Round(Clip(r\_y - r\_h / 2, min=0))
            \State y2 = Round(Clip(r\_y + r\_h / 2, min=H))
            \State input[:, :, x1:x2, y1:y2] = input\_s[:, :, x1:x2, y1:y2]
            \State lambda = 1 - (x2-x1)*(y2-y1)/(W*H) \Comment{Adjust lambda to the exact area ratio.}
            \State target = lambda * target + (1 - lambda) * target\_s \Comment{CutMix ends.}
        \EndIf
        \State output = model\_forward(input)
        \State loss = compute\_loss(output, target)
        \State model\_update()
    \EndFor
  \end{algorithmic}
  \label{alg:cutmix_algorithm}
\end{algorithm*}

We present the code-level description of CutMix algorithm in Algorithm~\ref{alg:cutmix_algorithm}.
{N, C, and K denote the size of minibatch, channel size of input image, and the number of classes.}
First, CutMix shuffles the order of the minibatch input and target along the first axis of the tensors.
And the lambda and the cropping region (x1,x2,y1,y2) are sampled. 
Then, we mix the input and input\_s by replacing the cropping region of input to the region of input\_s.
The target label is also mixed by interpolating method. 

Note that CutMix is easy to implement with few lines (from line $4$ to line $15$), so is very practical algorithm giving significant impact on a wide range of tasks.
\section{Weakly-supervised Object Localization}
\label{appendix:wsol}

We describe the training and evaluation procedure of weakly-supervised object localization in detail.

\noindent\textbf{Network modification: }
Basically weakly-supervised object localization (WSOL) has the same training strategy as image classification does. 
Training WSOL is starting from ImageNet-pretrained model. 
From the base network structures, VGG-16 and ResNet-50~\cite{resnet}, WSOL takes larger spatial size of feature map $14 \times 14$ whereas the original models has $7 \times 7$.
For VGG network, we utilize VGG-GAP, which is a modified VGG-16 introduced in~\cite{zhou2016CAM}.
For ResNet-50, we modified the final residual block (\texttt{layer4}) to have no stride ($=1$), which originally has stride $2$. 

Since the network is modified and the target dataset could be different from ImageNet~\cite{ImageNet}, the last fully-connected layer is randomly initialized with the final output dimension of $200$ and $1000$ for CUB200-2011~\cite{wah2011cub} and ImageNet, respectively.

\noindent\textbf{Input image transformation: }
For fair comparison, we used the same data augmentation strategy except Mixup, Cutout, and CutMix as the state-of-the-art WSOL methods do~\cite{singh2017hide,zhang2018ACOL}.
In training, the input image is resized to $256\times256$ size and randomly cropped $224 \times 224$ size images are used to train network.
In testing, the input image is resized to $256\times256$, cropped at center with $224\times224$ size and used to validate the network, which called single crop strategy.

\noindent\textbf{Estimating bounding box: }
We utilize class activation mapping (CAM)~\cite{zhou2016CAM} to estimate the bounding box of an object. 
First we compute CAM of an image, and next, we decide the foreground region of the image by binarizing the CAM with a specific threshold. 
The region with intensity over the threshold is set to 1, otherwise to 0. 
We use the threshold as a specific rate $\sigma$ of the maximum intensity of the CAM. 
We set $\sigma$ to $0.15$ for all our experiments. 
From the binarized foreground map, the tightest box which can cover the largest connected region in the foreground map is selected to the bounding box for WSOL.

\noindent\textbf{Evaluation metric: }
To measure the localization accuracy of models, we report top-1 localization accuracy (Loc), which is used for ImageNet localization challenge~\cite{ImageNet}. 
For top-1 localization accuracy, intersection-over-union (IoU) between the estimated bounding box and ground truth position is larger than $0.5$, and, at the same time, the estimated class label should be correct.
Otherwise, top-1 localization accuracy treats the estimation was wrong.

\subsection{CUB200-2011}
CUB-200-2011 dataset~\cite{wah2011cub} contains over 11 K images with 200 categories of birds.
We set the number of training epochs to $600$. 
For ResNet-50, the learning rate for the last fully-connected layer and the other were set to $0.01$ and $0.001$, respectively.
For VGG network, the learning rate for the last fully-connected layer and the other were set to $0.001$ and $0.0001$, respectively.
The learning rate is decaying by the factor of $0.1$ at every $150$ epochs.
We used SGD optimizer, and the minibatch size, momentum, weight decay were set to $32$, $0.9$, and $0.0001$. 

\subsection{ImageNet dataset}
ImageNet-1K~\cite{ImageNet} is a large-scale dataset for general objects consisting of 13 M training samples and 50 K validation samples.
We set the number of training epochs to $20$. 
The learning rate for the last fully-connected layer and the other were set to $0.1$ and $0.01$, respectively.
The learning rate is decaying by the factor of $0.1$ at every $6$ epochs.
We used SGD optimizer, and the minibatch size, momentum, weight decay were set to $256$, $0.9$, and $0.0001$.

\section{Transfer Learning to Object Detection}
\label{appendix:Detection}

We evaluate the models on the Pascal VOC 2007 detection benchmark~\cite{pascalvoc} with 5 K \texttt{test} images over 20 object categories.
For training, we use both VOC2007 and VOC2012 \texttt{trainval} (VOC07+12).

\noindent\textbf{Finetuning on SSD\footnote{https://github.com/amdegroot/ssd.pytorch}~\cite{liu2016ssd}: }
The input image is resized to $300 \times 300$ (SSD300) and we used the basic training strategy of the original paper such as data augmentation, prior boxes, and extra layers. 
Since the backbone network is changed from VGG16 to ResNet-50, the pooling location \texttt{conv4\_3} of VGG16 is modified to the output of \texttt{layer2} of ResNet-50.
For training, we set the batch size, learning rate, and training iterations to $32$, $0.001$, and  $120$ K, respectively.
The learning rate is decayed by the factor of $0.1$ at $80$ K and $100$ K iterations.

\noindent\textbf{Finetuning on Faster-RCNN\footnote{https://github.com/jwyang/faster-rcnn.pytorch}~\cite{fasterrcnn}: }
Faster-RCNN takes fully-convolutional structure, so we only modify the backbone from VGG16 to ResNet-50. 
The batch size, learning rate, training iterations are set to $8$, $0.01$, and $120$ K.
The learning rate is decayed by the factor of $0.1$ at $100$ K iterations.

\section{Transfer Learning to Image Captioning}
\label{appendix:captioning}

\begin{table*}[]
\centering
\begin{tabular}{@{}lccccccc@{}}
\toprule
                     & BLEU1 & BLEU2 & BLEU3 & BLEU4 & METEOR & ROUGE & CIDER \\ \midrule
ResNet-50 (Baseline) & 61.4  & 43.8  & 31.4  & 22.9  & 22.8   & 44.7  & 71.2  \\
ResNet-50 + Mixup     & 61.6  & 44.1  & 31.6  & 23.2  & 22.9   & 47.9  & 72.2  \\
ResNet-50 + Cutout    & 63.0  & 45.3  & 32.6  & 24.0  & 22.6   & 48.2  & 74.1  \\
ResNet-50 + CutMix    & \textbf{64.2}  & \textbf{46.3}  & \textbf{33.6}  & \textbf{24.9}  & \textbf{23.1}   & \textbf{49.0}  & \textbf{77.6}  \\ \bottomrule
\end{tabular}
\caption{Image captioning results on MS-COCO dataset.}
\label{tab:sup:imagecap}
\end{table*}

MS-COCO dataset~\cite{lin2014microsoft} contains $120$ K \texttt{trainval} images and $40$ K \texttt{test} images. 
From the base model NIC\footnote{https://github.com/stevehuanghe/image\_captioning}~\cite{vinyals2015show}, the backbone model is changed from GoogLeNet to ResNet-50. 
For training, we set batch size, learning rate, and training epochs to $20$, $0.001$, and $100$, respectively.
For evaluation, the beam size is set to $20$ for all the experiments.
Image captioning results with various metrics are shown in Table~\ref{tab:sup:imagecap}.

\section{Robustness and Uncertainty}
\label{appendix:robustness}

In this section, we describe the details of the experimental setting and evaluation methods.

\subsection{Robustness}

We evaluate the model robustness to adversarial perturbations, occlusion and in-between samples using ImageNet trained models. For the base models, we use ResNet-50 structure and follow the settings in Section 4.1.1.
For comparison, we use ResNet-50 trained without any additional regularization or augmentation techniques, ResNet-50 trained by Mixup strategy, ResNet-50 trained by Cutout strategy and ResNet-50 trained by our proposed CutMix strategy.

\noindent\textbf{Fast Gradient Sign Method (FGSM): }
We employ Fast Gradient Sign Method (FGSM) \cite{fgsm} to generate adversarial samples.
For the given image $x$, the ground truth label $y$ and the noise size $\epsilon$, FGSM generates an adversarial sample as the following
\begin{equation}
    \hat x = x + \epsilon ~\texttt{sign}\left(\nabla_{x} L(\theta, x, y)\right),
\end{equation}
where $L(\theta, x, y)$ denotes a loss function, for example, cross entropy function.
In our experiments, we set the noise scale $\epsilon = 8/255$.

\noindent\textbf{Occlusion: }
For the given hole size $s$, we make a hole with width and height equals to $s$ in the center of the image.
For center occluded samples, we zeroed-out inside of the hole and for boundary occluded samples, we zeroed-out outside of the hole.
In our experiments, we test the top-1 ImageNet validation accuracy of the models with varying hole size from $0$ to $224$.

\noindent\textbf{In-between class samples: }
To generate in-between class samples, we first sample $50,000$ pairs of images from the ImageNet validation set. For generating Mixup samples, we generate a sample $x$ from the selected pair $x_A$ and $x_B$ by $x = \lambda x_A + (1-\lambda) x_B$. We report the top-1 accuracy on the Mixup samples by varying $\lambda$ from $0$ to $1$.
To generate CutMix in-between samples, we employ the center mask instead of the random mask. We follow the hole generation process used in the occlusion experiments. We evaluate the top-1 accuracy on the CutMix samples by varing hole size $s$ from $0$ to $224$.

\begin{table*}[]
\centering
\begin{tabular}{@{}lccc|ccc@{}}
\toprule
Method   & TNR at TPR 95\% & AUROC & Detection Acc.  & TNR at TPR 95\% & AUROC & Detection Acc. \\ \midrule
         & \multicolumn{3}{c|}{TinyImageNet} & \multicolumn{3}{c}{TinyImageNet (resize)}       \\ \midrule
Baseline & 43.0 (0.0)    & 88.9 (0.0)   & 81.3 (0.0) & 29.8 (0.0)    & 84.2 (0.0)   & 77.0 (0.0)  \\
Mixup    & 22.6 \redp{-20.4}  & 71.6 \redp{-17.3} & 69.8 \redp{-11.5} & 12.3 \redp{-17.5}  & 56.8 \redp{-27.4} & 61.0 \redp{-16.0} \\
Cutout   & 30.5 \redp{-12.5}  & 85.6 \redp{-3.3}  & 79.0 \redp{-2.3}  & 22.0 \redp{-7.8}   & 82.8 \redp{-1.4}  & 77.1 \greenp{+0.1}  \\
CutMix   & \textbf{57.1 \greenp{+14.1}}   & \textbf{92.4 \greenp{+3.5}}   & \textbf{85.0 \greenp{+3.7}}  & \textbf{55.4 \greenp{+25.6}}   & \textbf{91.9 \greenp{+7.7}}   & \textbf{84.5 \greenp{+7.5}}  \\ \midrule
         & \multicolumn{3}{c|}{LSUN (crop)} & \multicolumn{3}{c}{LSUN (resize)}\\ \midrule
Baseline & 34.6 (0.0)    & 86.5 (0.0)   & 79.5 (0.0) & 34.3 (0.0)    & 86.4 (0.0)   & 79.0 (0.0)  \\
Mixup    & 22.9 \redp{-11.7}  & 76.3 \redp{-10.2} & 72.3 \redp{-7.2} & 13.0 \redp{-21.3}  & 59.0 \redp{-27.4} & 61.8 \redp{-17.2} \\
Cutout   & 33.2 \redp{-1.4}   & 85.7 \redp{-0.8}  & 78.5 \redp{-1.0} & 23.7 \redp{-10.6}  & 84.0 \redp{-2.4}  & 78.4 \redp{-0.6} \\
CutMix   & \textbf{47.6 \greenp{+13.0}}   & \textbf{90.3 \greenp{+3.8}}   & \textbf{82.8 \greenp{+3.3}} & \textbf{62.8 \greenp{+28.5}}   & \textbf{93.7 \greenp{+7.3}}   & \textbf{86.7 \greenp{+7.7}}  \\ \midrule
    & &   \multicolumn{3}{c}{iSUN} \\ \midrule
Baseline & & 32.0 (0.0)    & \multicolumn{1}{c}{85.1 (0.0}   & 77.8 (0.0)  \\
Mixup   & & 11.8 \redp{-20.2}  & \multicolumn{1}{c}{57.0 \redp{-28.1}} & 61.0 \redp{-16.8}  \\
Cutout  & & 22.2 \redp{-9.8}   & \multicolumn{1}{c}{82.8 \redp{-2.3}}  & 76.8 \redp{-1.0} \\
CutMix  & & \textbf{60.1 \greenp{+28.1}}   & \multicolumn{1}{c}{\textbf{93.0 \greenp{+7.9}}}   & \textbf{85.7 \greenp{+7.9}} \\ \midrule
         & \multicolumn{3}{c|}{Uniform} & \multicolumn{3}{c}{Gaussian} \\ \midrule
Baseline & 0.0 (0.0)     & 89.2 (0.0)   & 89.2 (0.0) & 10.4 (0.0)    & 90.7 (0.0)   & 89.9 (0.0)   \\
Mixup    & 0.0 (0.0)     & 0.8 \redp{-88.4}  & 50.0 \redp{-39.2} & 0.0 \redp{-10.4}   & 23.4 \redp{-67.3} & 50.5 \redp{-39.4} \\
Cutout   & 0.0 (0.0)     & 35.6 \redp{-53.6} & 59.1 \redp{-30.1} & 0.0 \redp{-10.4}   & 24.3 \redp{-66.4} & 50.0 \redp{-39.9} \\
CutMix   & \textbf{100.0 \greenp{+100.0}} & \textbf{99.8 \greenp{+10.6}}  & \textbf{99.7 \greenp{+10.5}}  & \textbf{100.0 \greenp{+89.6}}  & \textbf{99.7 \greenp{+9.0}}   & \textbf{99.0 \greenp{+9.1}}  \\ \bottomrule
\end{tabular}
\caption{Out-of-distribution (OOD) detection results on TinyImageNet, LSUN, iSUN, Gaussian noise and Uniform noise using CIFAR-100 trained models. All numbers are in percents; higher is better.}
\label{table:ood-sup}
\end{table*}

\subsection{Uncertainty}

Deep neural networks are often overconfident in their predictions. For example, deep neural networks produce high confidence number even for random noise \cite{hendrycks2016baseline}. One standard benchmark to evaluate the overconfidence of the network is Out-of-distribution (OOD) detection proposed by \cite{hendrycks2016baseline}. The authors proposed a threshold-baed detector which solves the binary classification task by classifying in-distribution and out-of-distribution using the prediction of the given network. Recently, a number of reserchs are proposed to enhance the performance of the baseline detector \cite{liang2017odin, lee2018simple} but in this paper, we follow only the baseline detector algorithm without any input enhancement and temperature scaling \cite{liang2017odin}.

\noindent\textbf{Setup: }
We compare the OOD detector performance using CIFAR-100 trained models described in Section 4.1.2. For comparison, we use PyramidNet-200 model without any regularization method, PyramidNet-200 model with Mixup, PyramidNet-200 model with Cutout and PyramidNet-200 model with our proposed CutMix.

\noindent\textbf{Evaluation Metrics and Out-of-distributions: }
In this work, we follow the experimental setting used in \cite{hendrycks2016baseline, liang2017odin}.
To measure the performance of the OOD detector, we report the true negative rate (TNR) at 95\% true positive rate (TPR), the area under the receiver operating characteristic curve (AUROC) and detection accuracy of each OOD detector.
We use seven datasets for out-of-distribution: TinyImageNet (crop), TinyImageNet (resize), LSUN \cite{yu2015lsun} (crop), LSUN (resize), iSUN, Uniform noise and Gaussian noise.

\noindent\textbf{Results: } 
We report OOD detector performance to seven OODs in Table~\ref{table:ood-sup}. Overall, CutMix outperforms baseline, Mixup and Cutout. Moreover, we find that even though Mixup and Cutout outperform the classification performance, Mixup and Cutout largely degenerate the baseline detector performance.
Especially, for Uniform noise and Gaussian noise, Mixup and Cutout seriously impair the baseline performance while CutMix dramatically improves the performance.
From the experiments, we observe that our proposed CutMix enhances the OOD detector performance while Mixup and Cutout produce more overconfident predictions to OOD samples than the baseline.

\end{document}